\documentclass{article}

\usepackage{arxiv}
\usepackage{csquotes}
\usepackage[usestackEOL]{stackengine}
\usepackage[utf8]{inputenc} 
\usepackage[T1]{fontenc}    
\usepackage{hyperref}       
\usepackage{url}            
\usepackage{booktabs}       
\usepackage{amsfonts}       
\usepackage{nicefrac}       
\usepackage{microtype}      

\usepackage{graphicx}
\usepackage{natbib}
\usepackage{doi}

\makeatletter
\renewcommand\scriptsize{\@setfontsize\scriptsize{7}{8}}
\renewcommand\tiny{\@setfontsize\tiny{5}{6}}
\renewcommand\small{\@setfontsize\small{12}{14}}
\renewcommand\normalsize{\@setfontsize\normalsize{12}{14}}
\renewcommand\large{\@setfontsize\large{14}{16}}
\renewcommand\Large{\@setfontsize\Large{14}{16}}
\renewcommand\LARGE{\@setfontsize\LARGE{14.4}{18}}
\renewcommand\huge{\@setfontsize\huge{20.74}{30}}
\renewcommand\Huge{\@setfontsize\Huge{24}{36}}
\makeatother

\title{Quantitative analysis of visual representation of sign elements in COVID-19 context}

\date{} 					

\author{ \href{https://orcid.org/0000-0002-7825-548X}{\includegraphics[scale=0.06]{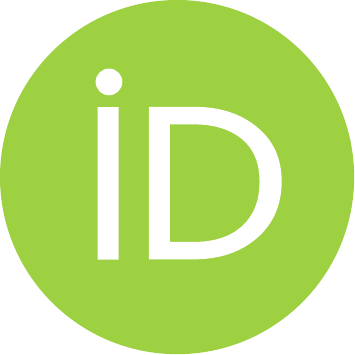}\hspace{1mm}Cano-Mart\'inez, Mar\'ia Jes\'us }\\
	Universidad Jaume I \\
	Castellon, Spain \\
	\And
	\href{https://orcid.org/0000-0002-5389-7590}{\includegraphics[scale=0.06]{orcid.pdf}\hspace{1mm}Carrasco, Miguel} \\
	Facultad de Ingeniería y Ciencias\\
	Universidad Adolfo Ibañez\\
	Santiago, Chile\\
	\And
	\hspace{1mm}Sandoval, Joaqu\'in\\
	Facultad de Ingeniería y Ciencias\\
	Universidad Adolfo Ibañez\\
	Santiago, Chile\\
    \And
	\href{https://orcid.org/0000-0002-9017-3196}{\includegraphics[scale=0.06]{orcid.pdf}\hspace{1mm}Gonz\'alez-Mart\'in, C\'esar } \\
	Universidad Adolfo Ibañez\\
	Santiago, Chile\\\thanks{corresponding author: \texttt{cgonzalez@unizar.es}}  \\
}

\renewcommand{\shorttitle}{Quantitative analysis of visual representation of sign elements in COVID-19 context}

\hypersetup{
pdftitle={Quantitative analysis of visual representation of sign elements in COVID-19 context},
pdfsubject={},
pdfauthor={Cano-Mart\i'nez, M.J., Carrasco, M., Sandoval, J. Gonzalez-Mar\'in, C.},
pdfkeywords={COVID-19, visual representation, computer analysis, the Covid Art Museum, Instagram},
}

\begin{document}
\strutlongstacks{T}
\maketitle

\begin{abstract}
	Representation is the way in which human beings re-present the reality of what is happening, both externally and internally. Thus, visual representation as a means of communication uses elements to build a narrative, just as spoken and written language do. We propose using computer analysis to perform a quantitative analysis of the elements used in the visual creations that have been produced in reference to the epidemic, using the images compiled in The Covid Art Museum's Instagram account to analyze the different elements used to represent subjective experiences with regard to a global event.  This process has been carried out with techniques based on machine learning to detect objects in the images so that the algorithm can be capable of learning and detecting the objects contained in each study image. This research reveals that the elements that are repeated in images to create narratives and the relations of association that are established in the sample, concluding that, despite the subjectivity that all creation entails, there are certain parameters of shared and reduced decisions when it comes to selecting objects to be included in visual representations

\end{abstract}


\newpage
\section{Introduction}

\begin{displayquote}
«(…) the best or most healthy way to respond to illness is through narrative»
\citep[p.116]{woods_beyond_2012}
\end{displayquote}

We live surrounded by images. Thus, the representational baggage that things contain is not in fact overcome through discourse, but also through the images involved in every construct. In this sense, we have forgotten that «every image has at least two components: the reality it is reproducing and the meaning of that represented reality» \citep[p.46]{aparici_imagen_2009}. Beyond the merely mundane, human beings have developed a sophisticated way of creating images, which far from presenting reality, represent it, assigning it a story, a function. Thus, the image acts as a testimony to events, as liberator of expressions or emotions, or as a merely persuasive, a weapon capable of inciting action and/or consumption. 

Jovchelovitch and Routledge (\citeyear{jovchelovitch_knowledge_2019}) shows us in his research how representation is the basis of every knowledge system and the origin of understanding. Representation allows us to develop and implement the social constructionist approach \citep{jhally_representation_1997} by favoring our understanding of the world and the relationships among members of a community.

More specific studies on the function of the image, in addition to the ontological role it plays \citep{banks_rethinking_1999,chaplin_sociology_1996,grady_scope_1996,ruby_visual_1996}, seek to materialize its effects at the social \citep{harper_visual_2012}, or individual/psychological \citep{balcetis_social_2010,reavey_visual_2011} levels. Thus, on the sociological level, Grady (\citeyear{grady_scope_1996}) points out the following items that images contribute to social studies and construction: «The first is Seeing, or the study of role of sight and vision in the construction of social organization and meaning. The second is Iconic Communication, or the study of how spontaneous and deliberate construction of images and imagery communicate information and can be used to manage relationships in society. The third area I have termed Doing Sociology Visually and is concerned with how techniques of producing and decoding images can be used to empirically investigate social organization, cultural meaning and psychological processes» (p. 10).
These aspects, which are aimed at the influence images have over social constructs, are reinforced in the compendium of qualitative image studies by Reavey (\citeyear{reavey_visual_2011}), who argues that the visual is an integral part of the way culture operates and especially now, thanks to the mass media. 

As a complement to this influence of the image under a sociological approach, at the psychological level visual imagery is appreciated for its «(...) impact on how people experience the world they live in» \citep[XVI]{reavey_visual_2011}. This is why it is hard to ignore the complexity of the visual elements that mediate between personal experience and context. Therefore, when we take seriously how people's experiences are created and the contexts in which they emerge, it becomes hard to ignore the rich complex of visual media through which experiences are made reality.

Thus, when faced with a global event that has altered and modified our way of life, limiting our freedom and our individuality, as has been the case with the COVID-19 pandemic, this research aims to detect which denotative elements people use to express their experiences in this context through their visual creations and to detect whether common patterns and ideas emerge among the creators.

\newpage
\subsection{The power of the image: effects of representation}
\begin{displayquote}
«Representation is an essential part of the process by which meaning is produced and exchanged between members of a culture. It does involve the use of language, of signs and images which stand for or represent things» \citep[p.15]{jhally_representation_1997}.
\end{displayquote}

Representation entails a system that combines the codes that we are capable of interpreting through learning and perception on the one hand, and the construction of signs on the other, which —in our case— visually represent concepts through «(...) ways of organizing, clustering, arranging and classifying concepts, and of establishing complex relations between them» \citep[p.17]{jhally_representation_1997}. A language.

Thus, visual representation is a process of social, cultural and ideological mediation \citep{desai_imaging_2000,knochel_assembling_2013}, selection, classification and deconcontextualization of elements with information from reality, with which a meaning is constructed through the structuring of these representation elements. This process fosters the transmission of an informative and/or emotional message based on individual or collective experience, which will ultimately be broadcast to the world in a language that must be shared by a community. 

We speak of construction because a representation cannot reflect reality, as we all have different experiences and interpretations of 'reality'. In the case of the pandemic, we have all shared the same framework, but our experiences and interpretations of events, of reality, show subjective variations in their representation/description, meaning that the image created as artistic practice «(...) has come to be replaced by presentation, by a form of presenting reality more than representing it, while art, especially contemporary art, flees the mimesis to present a new reality» \citep[p.104]{cano_escondido_2018}. Thus, when it comes to the pandemic we cannot speak of representations, but rather of presentations and dynamic constructs.

Some research assumes «(...) that images are not surfaces of universal messages whereby a common sense interpretation becomes apparent to all who look on. Instead, images serve as filters of how we see and are seen in the world, tinting our perceptions about our peers, building a bricolage of our understanding and assumptions about the other, and looking back at us to inform our self-concept» \citep[p.17]{knochel_assembling_2013}. However, in the pandemic context we are in, can the message of a subjective representation of experience be universal?

\subsection{Visual semiotics: elements for a language of representation}
The subjective world created by visual representation is perceived with material elements «(...) whose 'shape' includes all the meaning of the work» \citep[p.433]{burgess_self-representation_2018}, which constitutes a visual message. Following Visual Literacy (VL) theory, which defines itself as an interdisciplinary, multidisciplinary and multidimensional discipline for the development of skills to interpret and create visual messages through the use and manipulation of cultural elements \citep{burgess_self-representation_2018}, it is akin to a verbal and literary language, and as a language, the visual requires skills, competencies and learning to properly convey and perceive the message: «(...) VL involves cognitive functions such as critical viewing and thinking, imaging, visualizing, inferring as well as constructing meaning; but also communicating as well as evoking feelings and attitudes» \citep[p.267]{cohn_visual_2020}.

As a language, visual representation is composed of graphic elements such as shapes, lines, colors, textures, etc.—letters in the linguistic context—primitive components that make up more complex elements (e.g. a face) - the word in traditional language. This «(...) involves an interaction between three primary structures, similar to the parallel architecture of linguistic systems (...): A meaning expressed by a modality (here: visual-graphic marks), which is organized using combinatorial grammatical structures»(Jackendoff cited in \citep[p.267]{cohn_visual_2020}.

These complex forms fulfill a function and become significant, depending on the cultural context; they contain meaning (concept) in themselves and the sign is established by combining them. The combination of signs in an image determines the message through a spatial structure of relationships. In the case of visual language, their correct perception and understanding will be determined by the degree of iconicity. 

The visual features of images can be divided into general features (color, shade, texture, etc.) of domain-specific features like human faces \citep{sethi_mining_2001}. This division is similar in semantics, which establishes two levels: Low-Level description representation, which are the formal characteristics of the image: color, light, textures, etc. dependent on the concept that one seeks to transmit \citep{hutchison_semantic_2005}; and the highest level, which is the meaning (the concept) \citep{liu_survey_2007}, established by the semantic relations between elements.

In our research we analyze signs at the denotative - descriptive - level, contents and representations in the images developed by users on the Internet to identify the signs, and combinations used by users in their visual representations based on the pandemic. Thus, like written or spoken language, we can discover the most-used words and phrases and reach a more connotative level of the expressions of shared feelings during the pandemic on the part of a group of people at the global level and compiled on the social network Instagram.

\begin{displayquote}
«Connotations are common associations connected to a sign, not private associations that only one individual might have, but associations and references that are shared by larger cultures or groups» \citep[p.433]{burgess_self-representation_2018}. 
\end{displayquote}

\subsection{Health humanities. Incorporation of a visual narrative as discursive element of the disease}
Health humanities is an interdisciplinary field of study that seeks to incorporate a humanistic perspective to health and healthcare. Humanities (communications, cultural studies and linguistics, history, literature, philosophy, theology and religious studies and arts) are indispensable to human healthcare as they teach us what it means to be human. They teach us about the human condition, about human suffering and healing, and about human well-being and flourishing. 

In a systematic review, the authors Costa, Kangasjarvi and Charise (\citeyear{costa_beyond_2020}) provide us with the following data on the benefits of applying the humanities to the field of health: «(...) proponents of 'medical humanities' have argued that by encouraging attention to less overtly clinical aspects of healthcare, humanities-engaged clinical education helps build social and relational skills, often by fostering empathy and moral development; compassion; self-reflection; interpersonal skills; perspective-taking; openness to otherness; critical reflection; and tolerance for ambiguity.» (p. 1204). 

In reference to representation, and specifically social representation, Moscovici (\citeyear{moscovic_introduccion_1985}) argues there are three fundamental types: emancipated, controversial and hegemonic. These types of representation are also part of the metaphorical developments of different social constructs. In the case we are concerned with, it could be the social representation of the COVID-19 pandemic. Art can be situated in each of these, exercising different functions in the construction of a narrative of events, in addition to the dissemination and empowerment of some discourse over others. An example of this, regarding the second type described by Moscovici (hegemonic social representations), was presented by Alfonso Baya (\citeyear{baya_imaginario_2015}) in his thesis on the imagery of AIDS and contributions from Art, where he points out the main hegemonic representations: «The visual elements that the pictorial images have offered of disease over the course of history have been allegories of the disease in itself, such as JF Bocanegra's 'Allegory of the Plague' in the 1660s. The figure of the doctor visiting the sick, such as "Doctor's Visit" by J Steen (1625-79), or the terminally ill in bed. Other representations are directly inclined toward the representation of death, either as an allegory or as a human still life without action, movement or life» (p. 71). This leads us to wonder whether the narrative that is being created of the pandemic is putting forward certain hegemonic constructs of the same, above the display of the diversity of existing social realities; whether it is oriented toward the dissemination of images more aimed at controversy and the confrontation of opinions on the disease and its handling in different aspects, or whether it is fulfilling an emancipating function instead, where art is presented as a solution to the circumstances that the pandemic places us in

\subsection{Computer analysis of visual representation}
From a behavioral perspective, studies indicate that visual perception of art occurs in a differentiated way among groups of artists and non-artists \citep{solso_cognitive_2000, zeki_inner_2003}. Human vision is not just perceived in terms of the visual aspect of the work of art, but rather there is an understanding that takes the artist beyond the work and even beyond the aesthetic terms that might be represented in it \citep{ramachandran_science_1999}. This means that artists not only paint through their eyes, but rather through an understanding and interpretation of what is happening in their brains \citep{goguen_art_1999, quiroga_mendez_dar_2016, zeki_inner_2003}.

Thus, the question emerges as to whether it is possible for a computer to capture the nature of a painting. That is, what we experience in human form as the capacity for imagination and connection with art. The first works aimed at understanding art from the field of computer sciences were developed over two decades ago, thanks to the initial work of Taylor (\citeyear{taylor_fractal_1999}) to authenticate the works of art by Jackson Pollock. For their part, Widjaja et al. (\citeyear{widjaja_identifying_2003}) proposed the combination of multiple classifiers of four artists through skin pattern analysis. Li and Wang (\citeyear{li_studying_2004}) developed an algorithm for the automatic identification of Chinese ink paintings through wavelet analysis and MHMM (Multiresolution Hidden Markov Models) tools. Johnson et. al. (\citeyear{johnson_image_2008}) proposed an algorithm that uses a set of wavelet filters on different scales to extract patterns from Van Gogh's and other paintings (not Van Gogh). Shamir et al. (\citeyear{shamir_impressionism_2010}) proposed an automatic recognition system that differentiates the works of three types of art school: impressionism, abstract expressionism, and surrealism. For this they used a set of algorithms to extract characteristics based on textures, statistical moments, Gabor filters and the analysis of edges, among others. As the literature describes, all the techniques prior to 2012 use the manual extraction of low-level characteristics according to chromatic, geometric, or statistical descriptors, which in general perform poorly as they do not extract the underlying information from the styles contained in the images.

Though neural networks \citep{zhang_neural_2000} have been present for more than 60 years, it was not until the methodology proposed by Krizhevsky et al. (\citeyear{krizhevsky_imagenet_2017}) that it marked a paradigm shift by implementing a convolutional neural network (CNN) with a more complex and operational architecture, which allowed automatic extraction of patterns from images without requiring the work of manual characteristic extraction. This last tool has been a major revolution, as CNNs have made it possible to extract mid-level properties present in images due to the internal process of the network itself. In the years that followed, exhaustive analyses of different sets of data have shown the great flexibility and power of extracting patterns from images, which can even be transferred even to other domains \citep{oquab_learning_2014, razavian_cnn_2014,wang_deep_2018,hua_detecting_2016}.

\begin{figure}
	\centering
	\includegraphics[width=\textwidth]{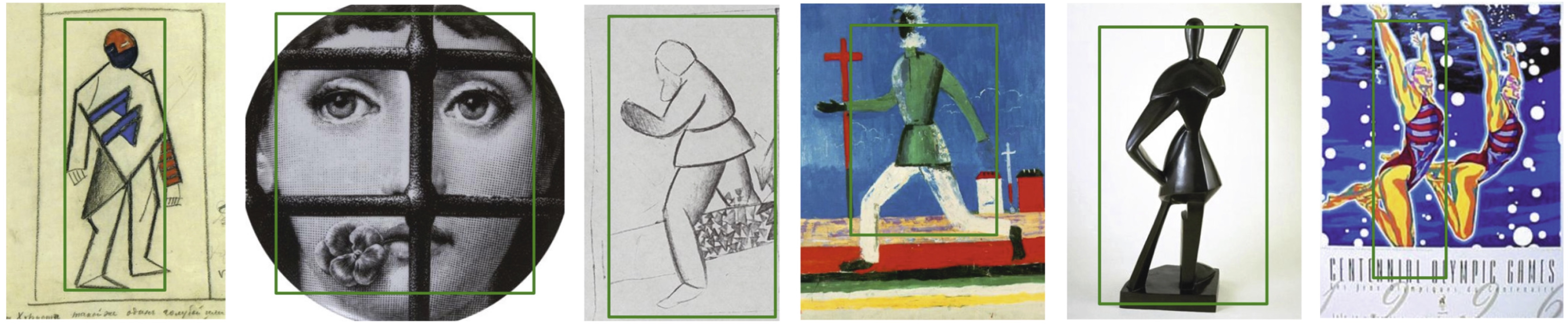}
	\caption{Semantic identification in different styles of painting on which the algorithm identifies the spatial location. Green box on image \citep{hua_detecting_2016}.}
	\label{fig:fig1}
\end{figure}

Currently extensive databases have been developed that are composed of thousands or millions of images of digital photographs with their corresponding objects classified (i.e., SUN, PascalVOC, ImageNet, Google Image Database), which have allowed major progress to be made in identifying, recognizing, and classifying images and their objects \citep{russakovsky_imagenet_2015}. This does not normally occur in artistic paintings where different styles are present and these change over time, in addition to the limited universe of artistic images compared to digital images like photographs of everyday objects \citep{crowley_state_2014, garcia_covid19_nodate, hong_art_2019, ren_faster_2017}. This leads to substantial differences in the statistics associated with the objects represented in the images. However, thanks to the development of CNN in digital photographs, it has been possible to transfer this knowledge to the area of art and paintings, in this way overcoming the adaptation of domain \citep{kulis_what_2011, wang_deep_2018}. An example of this is found in Karayev (\citeyear{karayev_recognizing_2014}) work, which is aimed at differentiating the different esthetics not just of art, but also of digital images where different photographic techniques, compositions styles, moods, genders, and types of scenes can be distinguished. The same is the case with the work by Crowley and Zisserman (\citeyear{crowley_state_2014}), who managed to detect and align objects in paintings based on their learning of the same using natural images (i.e., not paintings). In the same way, Westlake (\citeyear{hua_detecting_2016}) makes progress in the study of artistic images, especially with the creation of a database of specialized in people with over 43 styles of painting. He emphasizes the use of learning transfer from the ImageNet network, modifying the output network. The study reveals the complexity that an identification algorithm faces in detecting a given study subject. In the case of detecting people, this can be done with multiple art styles and even so, the algorithm can identify their spatial location within the image (see Fig.~\ref{fig:fig1}). This is a complex task given the countless types of configuration, form, pose, size, and abstraction that exist in the artists' interpretations.

In recent years, identification tools based on Deep Learning architectures \citep{sandoval_two-stage_2019, tan_ceci_2016} have evolved and significantly improved accuracy, speed and performance on issues related to identification of regions, known as region multi-state detectors. This category includes the following frameworks: SPP-net (spatial pyramid pooling)~\citep{he_spatial_2014}, R-FCN (region based fully convolutional network) (region-based fully convolutional network)~\citep{dai_r-fcn_2016}, FPN (Feature pyramid networks)~\citep{lin_feature_2017}, Fast R-CNN \citep{ren_faster_2017}, Faster R-CNN~\citep{ren_faster_2017}, Mask R-CNN \citep{he_mask_2017}and R-CNN~\citep{zhao_object_2019}. Despite the wide variety of identification and classification tools, one of the architectures that has performed the best, especially with abstract images, has been the "You Only Look Once” (YOLO) network \citep{hua_detecting_2016}. YOLO, proposed by Redmon et al.~(\citeyear{redmon_you_2016}), is a simplified network compared to previous architectures, requiring only reducing the size of the entire image and processing it directly over the network. Other networks usually require a preliminary process that consists in scrolling over the image with a sliding window and then using an adjustment process to find objects of interest along with adjusting the boundaries of each object found. In addition, YOLO processes the image by dividing it into a rectangular grid and then searching for and grouping quadrants together as the likelihood of finding an object increases. This process works globally, thus increasing the general performance in object detection. This means that the probabilities of the objects found in the image are simultaneously predicted and adjusted. For YOLO to operate, there is a training stage that the developers themselves have provided for, though it can be trained for a specific set of images \citep{buric_adapting_2019,dutta_cnn_2019, ju_application_2019}. YOLOv4 is currently the most accurate and fastest network among current identification tools \citep{bochkovskiy_yolov4_2020, sumit_object_2020-1} and was used as the key object identification tool in this research.

\section{Materials and methods}
This article analyzes visual representations arising from the COVID-19 pandemic and which suppose an approach from a perspective governed by a critical, reflexive vision capable of bringing our most immediate reality closer to us. Now this disease, close and threatening, brings up the occasionally forgotten idea of people's vulnerability and fragility, seeking through the compilation of the different works reviewed to value their discourse and the existence of different approaches to the situation generated by the pandemic, in addition to the reflections that it has prompted. A new disease - without a visual baggage or background like we have been able to see with cancer, HIV or mental disease in the history of contemporary art, which have been represented and loaded with significance - and which has nevertheless produced a huge number of images around it \citep{garden_disability_2010, garden_who_2015}.

\begin{figure}
	\centering
	\includegraphics[width=\textwidth]{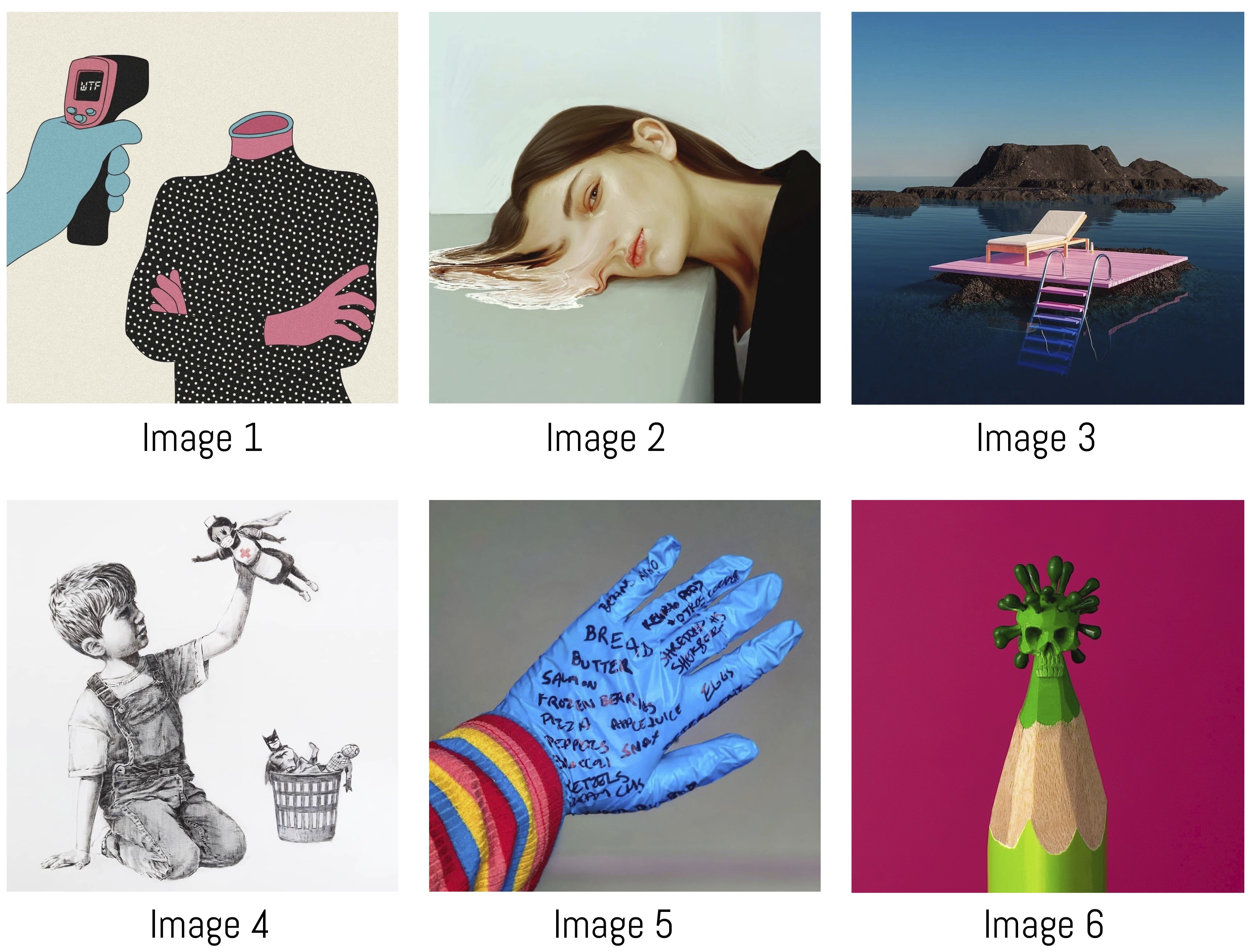}
	\caption{Sample images from Instagram's Covid Art Museum account (Image 1: Hojin Kang, @hojinkangdotcom; Image 2: @Bix.rex.1; Image 3: @marciorodriguezphoto; Image 4: @rebordacao, Rebordação Embroidery; Image 5: Orane Tasky @oranetasky; Image 6: Pauz Peralta, Title: The Fridge, @notpauz).}
	\label{fig:fig2}
\end{figure}

The compilation of images created in the context of the pandemic for this research has been done using the Instagram profile entitled The Covid Art Museum (CAM),  created as an initiative to collect the visual expressions and representations produced by Internet users where they reflect on their experiences during the health emergency \citep{wullschlager_how_2021} (see samples in Fig.\ref{fig:fig2}). The period for the mass-download of images ranges from the date that the profile was created (19 March 2020) to 4 August 2020, with a total of 927 images obtained.

Since the start of the pandemic, proposals like the Covid Art Museum have been repeated in the digital plane with other Instagram accounts, concrete proposals by galleries and museums, such as the Alameda Art Center in Santiago  or the Reina Sofia Museum in Spain,  in addition to virtual exhibits on the problem, where we can highlight the MURAL proposal by the Spanish National Museum of Anthropology. 

In the qualitative analysis of the works, it is important to recognize whether the discourses on COVID-19 and its consequences and side-effects have, as Susan Sontag (2017) analyzes in her work, left a series of associated stigmatizing taboos and/or metaphors, or whether they are more oriented toward resilience. Our work picks up on this idea and, as Jesús Martínez (\citeyear{martinezoliva_mirando_2013}) underscores regarding the reactivation of artistic discourse through disease and the messages and discourses that build it, examine the concrete contributions that the artistic experience itself brings to the experience of the disease, dealing with it and its visibility, in a moment in which even its very way of conceiving and promoting itself is changing.

\subsection{Computer analysis}
Current techniques based on Deep Learning algorithms generally have a limited number of objects that the network has been trained for. These networks generally work with objects or representations from the physical world and not necessarily abstract objects like those commonly found in art. Though there is a growing body of research related to the identification of elements and/or representations in artistic expressions, it is always done based on a limited and defined set. 
  The images used in our study represent a mix of physical objects and/or abstract representations. Analyzing this type of images is complex, especially due to the process of abstraction that the artist defines in his or her composition.The methodology proposed seeks to find the relationship between the objects in each image to find a general ontology that brings together the concepts contained in each image and how they are related to the current context of the COVID-19 pandemic. To facilitate this search, there was a need to redefine the classes as a function of the objects they contain, thus facilitating the networks learning process

The analysis proposal is made up of four phases, which we detail below (Fig.\ref{fig:fig3}). The first phase consists in preparing the data. In this phase each of the objects to be detected are manually identified in the following categories:

\begin{figure}
	\centering
	\includegraphics[width=\textwidth]{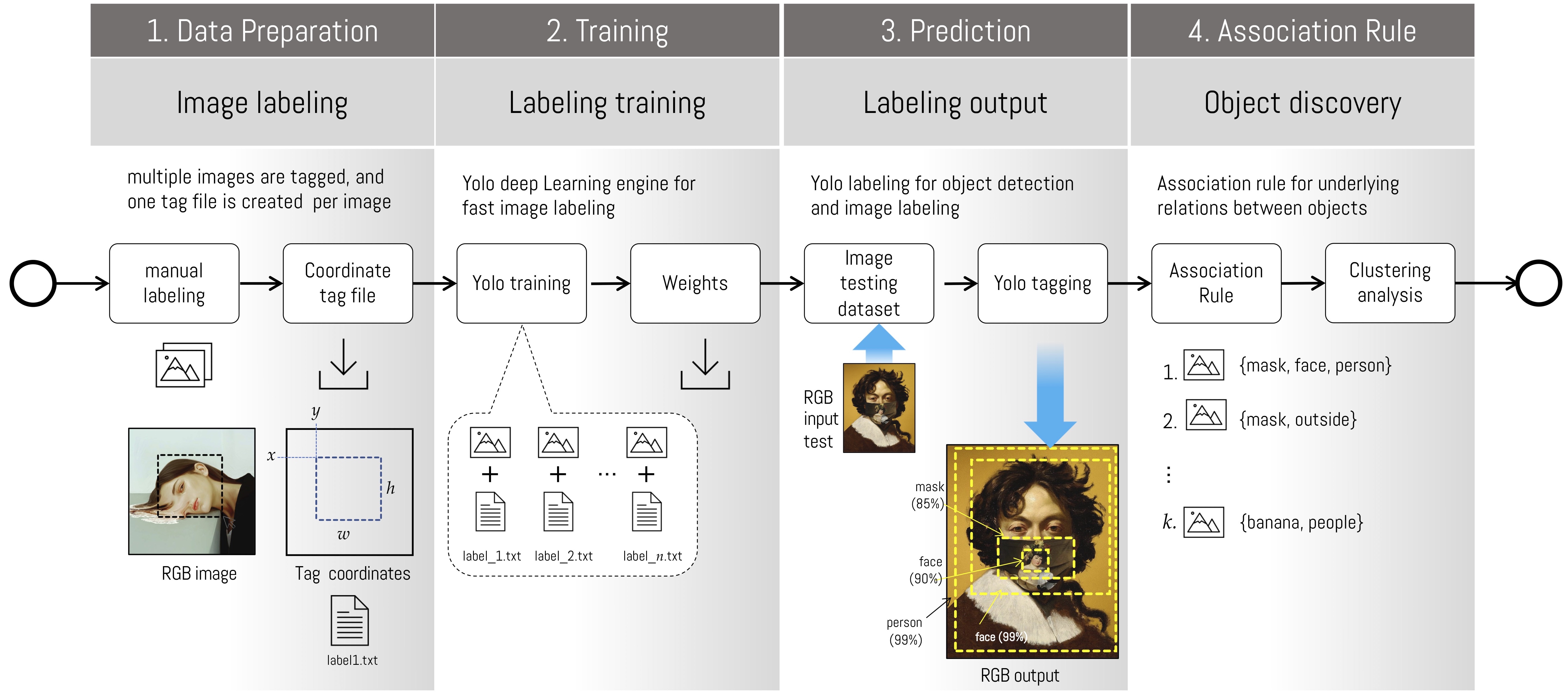}
	\caption{Sample images from Instagram's Covid Art Museum account (Image 1: Hojin Kang, @hojinkangdotcom; Image 2: @Bix.rex.1; Image 3: @marciorodriguezphoto; Image 4: @rebordacao, Rebordação Embroidery; Image 5: Orane Tasky @oranetasky; Image 6: Pauz Peralta, Title: The Fridge, @notpauz).}
	\label{fig:fig3}
\end{figure}

\newpage
\begin{quote}
Categories = { dog, person, cat, TV, car, outside, plants, books, window inside, bird, closed curtains, bicycle, plate, banana, ketchup, cup, hanging clothes, house, rubber gloves, garlic, vase, mask, shopping cart, groceries, toilet paper, abstract mask, handwashing, faucet, virus, nudity, kiss, phone, flower, fruits, wine, bathtub, bread, toilet paper pile, sanitizer, snacks, game controller, empty toilet, paper roll, soap, laptop, knife, bed, balcony, disinfecting wipes, bleach, soap bar, pen, syringe, motorcycle, thermometer, crowd, city landscape, street, buildings, bathroom, supermarket, soap dispenser, dish soap, videochat}
\end{quote}

The process of choosing the categories was inductive. It bega with the compilation of objects contained in the images and associating them with more general concepts. We defined 64 categories for our study, broad enough to detect a large proportion of the elements contained in the images. 

A rectangular region is identified for each object and associated with one of the categories listed above. In some cases, due to the large number of objects contained in the same image, we have manually identified 60\% of the objects contained in it. That is, if an image contains more versions of the same object, then only 60\% of these are tagged. 

In the second phase (training), we used a YOLOv4 convolutional neural network. The data was divided into three groups in this phase: training (70\%), validation (20\%), and test (10\%). In this way, all the previously identified objects are trained in the network to be automatically classified in subsequent steps. As results, we put forward a set of network training weights specific to the set of images of the problem.  In the third phase the network is assessed on the same set of training images. Though we know that normally we seek to find objects in images that have not previously been trained, it is important to note that not all the objects were tagged in many of the training set's images. In this way we increased the number of objects detected in the training images. In contrast, we did not use YOLO as a prediction tool, but rather as a complementary tool for the object identification process.   

Lastly, the fourth phase uses all the objects detected in each image as a knowledge base. To find a relationship among these and their relationship with different images, we used an association rule algorithm \citep{agrawal_fast_1996} that allows the ontology in the images to be defined. In this way, in addition to finding the frequency of the objects' appearance in each image, the rules of association allow us to determine the underlying relationship in objects that do not initially seem to be related but which are, through the search of patterns that group the relationship together. To facilitate the visual analysis, we used the graphic representation of the rules of association, in which each node represents an object from the set of study images and the arrow indicates the relationship between antecedent items (LHS) and consequent items (RHS). For example, if an object {buildings, street} repeatedly appears in an image, then it implies a certain level of probability that the object {outside} will also be present in the image. In this way, the algorithm can build a relationship according to two indicators: Support and confidence, where support determines the percentage of relationships that appear in the dataset and Confidence is the frequency with which consequent items appear along with antecedent ones

\section{Results}

\begin{figure}
	\centering
	\includegraphics[width=\textwidth]{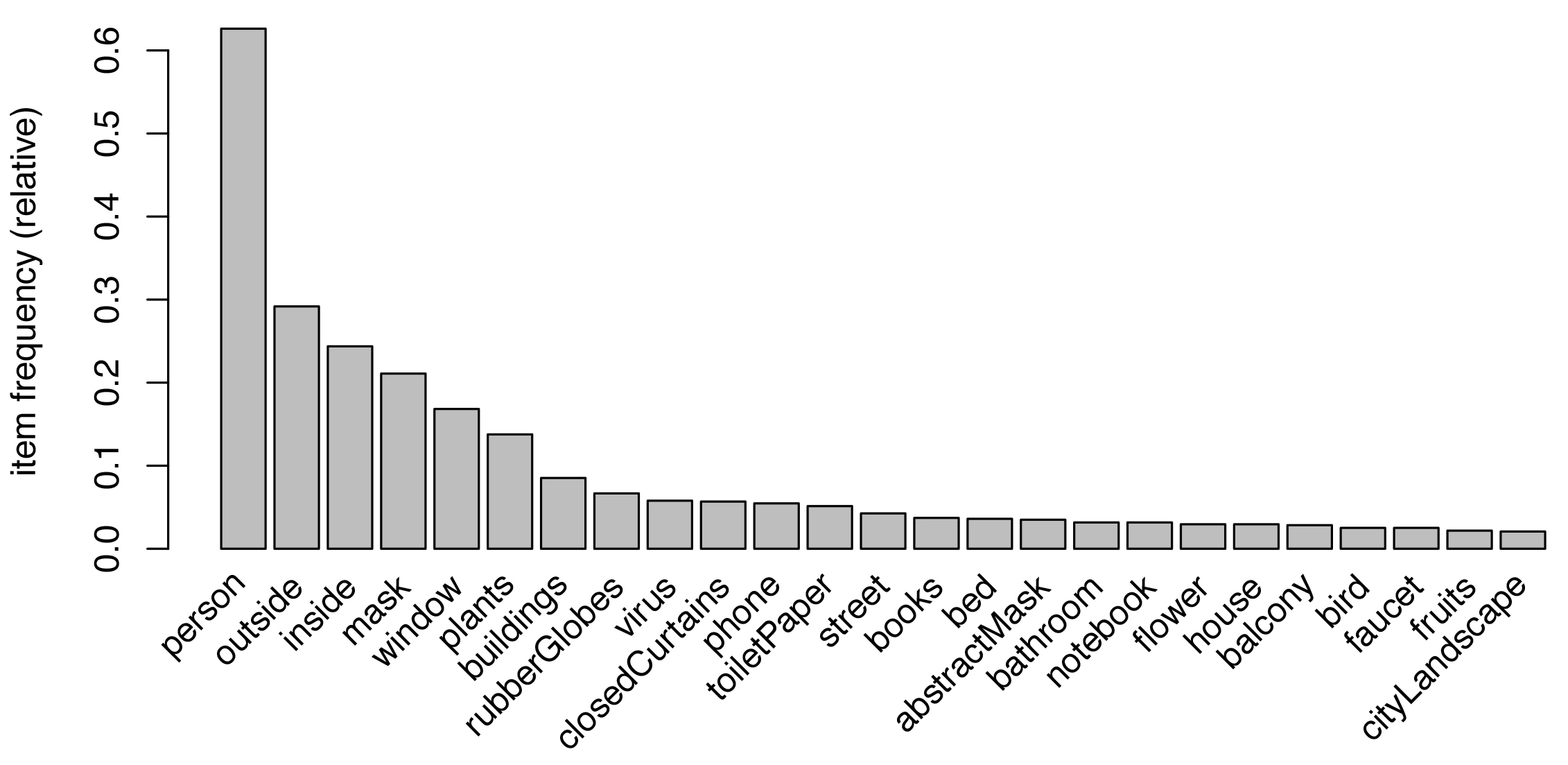}
	\caption{Process of preparing and identifying objects and their relationships.}
	\label{fig:fig4}
\end{figure}

The sample analysis reveals that the most common elements used to construct narratives are people, an outside view, an interior scene, face masks, windows, vegetable elements and buildings (see Fig.~\ref{fig:fig4}). However, frequency-based analysis only allows identifying the objects that appear most frequently in the set of images, but it does not identify the combinations between the elements represented, which constitute the decision-making to build the message, like the sum of words in a sentence.
This limitation has been resolved with a priori analysis, which allows finding the relationships between multiple objects, as shown in Table 1, ordered according to confidence.

\begin{table}
	\caption{Result of the a priori algorithm for the first 20 rules of association ordered according to confidence: LHS (antecedent), RHS (consequent)}
	\centering
	\scalebox{0.8}{
	\begin{tabular}{p{0.5cm}lllll}
ID  & \Longstack{LHS\\(ANTECEDENT})& \Longstack{RHS\\(CONSEQUENT)}& SUPPORT& CONFIDENCE& LIFT\\
\toprule
1	&{kiss}	&{person}	& 0.0120	&1.0000	&1.597\\
2	&{nudity}	&{person}	&0.0175	&1.0000	&1.597\\
3	&{kiss,mask}	&{person}	&0.0109	&1.0000	&1.597\\
4	&{inside,nudity}	&{person}	&0.0131	&1.0000	&1.597\\
5	&{books,window}	&{person}	&0.0131	&1.0000	&1.597\\
6	&{buildings,street}	&{outside}	&0.0153	&1.0000	&3.427\\
7	&{cityLandscape,outside,person}	&{buildings}	&0.0109	&1.0000	&11.731\\
8	&{outside,person,rubberGloves}	&{mask}	&0.0109	&1.0000	&4.741\\
9	&{buildings,closedCurtains,outside}	&{window}	&0.0186	&1.0000	&5.942\\
10	&{buildings,closedCurtains,person}	&{window}	&0.0131	&1.0000	&5.942\\
11	&{buildings,closedCurtains,outside,person}	&{window}	&0.0120	&1.0000	&5.942\\
12	&{closedCurtains,outside}	&{window}	&0.0284	&0.9630	&5.722\\
13	&{buildings,closedCurtains}	&{window}	&0.0197	&0.9474	&5.629\\
14	&{closedCurtains,outside,person}	&{window}	&0.0197	&0.9474	&5.629\\
15	&{buildings,closedCurtains,window}	&{outside}	&0.0186	&0.9444	&3.237\\
16	&{mask,window}	&{person}	&0.0175	&0.9412	&1.503\\
17	&{abstractMask}	&{person}	&0.0328	&0.9375	&1.497\\
18 &{closedCurtains,inside}	&{person}	&0.0164	&	&1.497\\
19	&{inside,outside}	&{person}	&0.0164	&	&1.497\\
20	&{cityLandscape,person}	&{buildings}&0.0131	&	&10.828\\
	\end{tabular}}
	\label{tab:table}
\end{table}

\begin{figure}
	\centering
	\includegraphics[width=\textwidth]{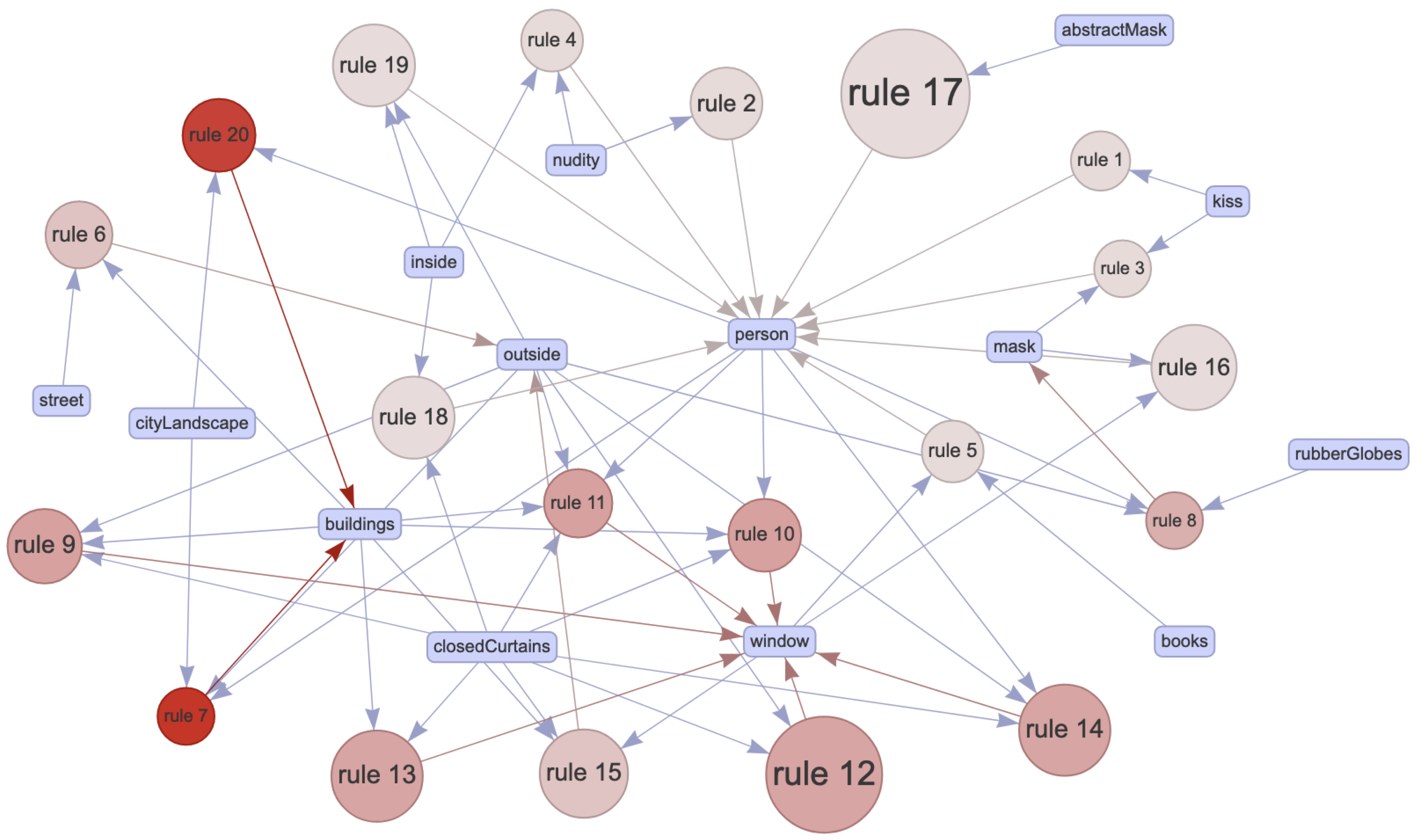}
	\caption{A priori relationship between the objects found within the same image.}
	\label{fig:fig5}
\end{figure}

\begin{figure}
	\centering
	\includegraphics[width=\textwidth]{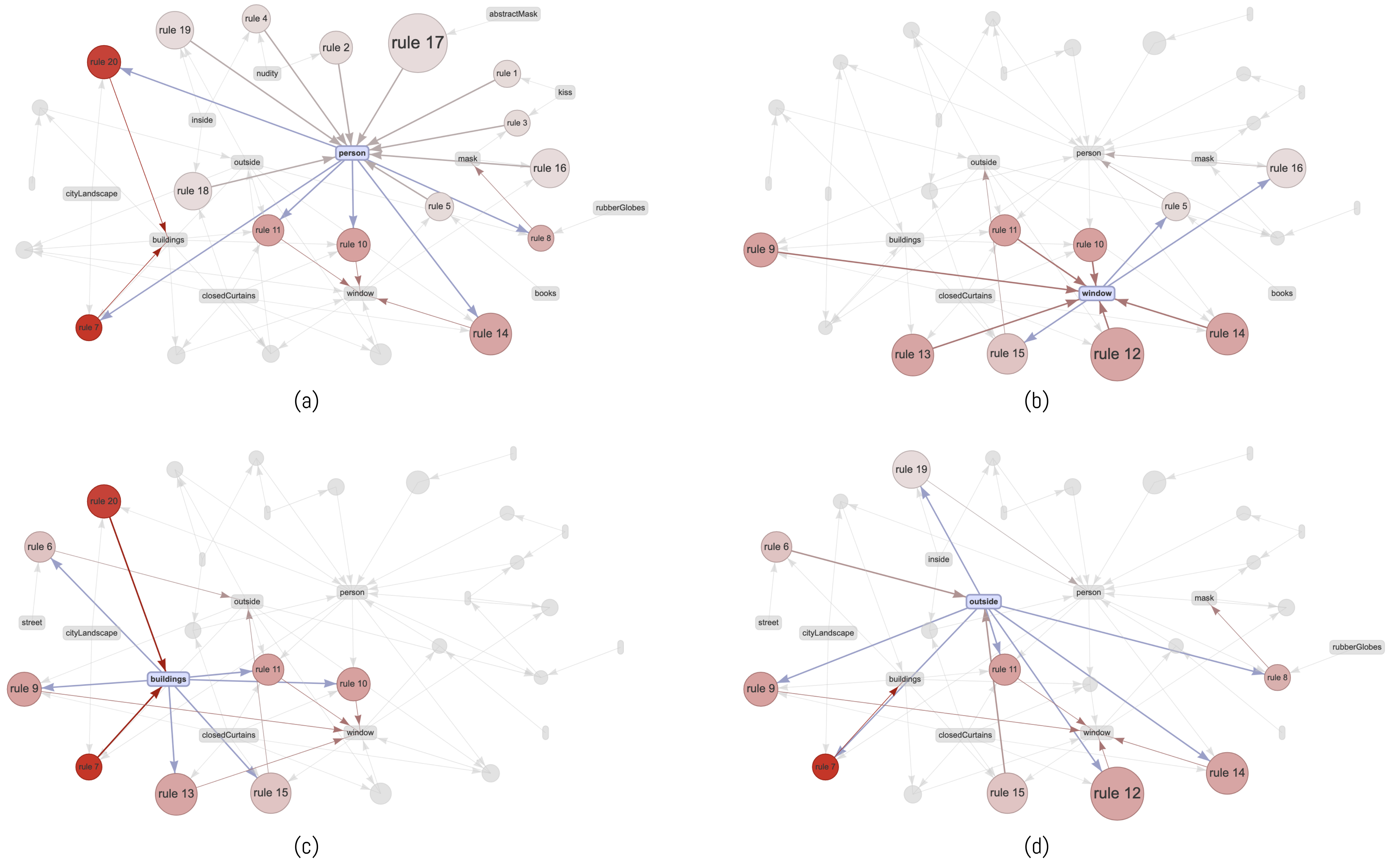}
	\caption{Four main antecedents with the largest number of rules of association between antecedents and consequents.}
	\label{fig:fig6}
\end{figure}

\begin{figure}
	\centering
	\includegraphics[width=\textwidth]{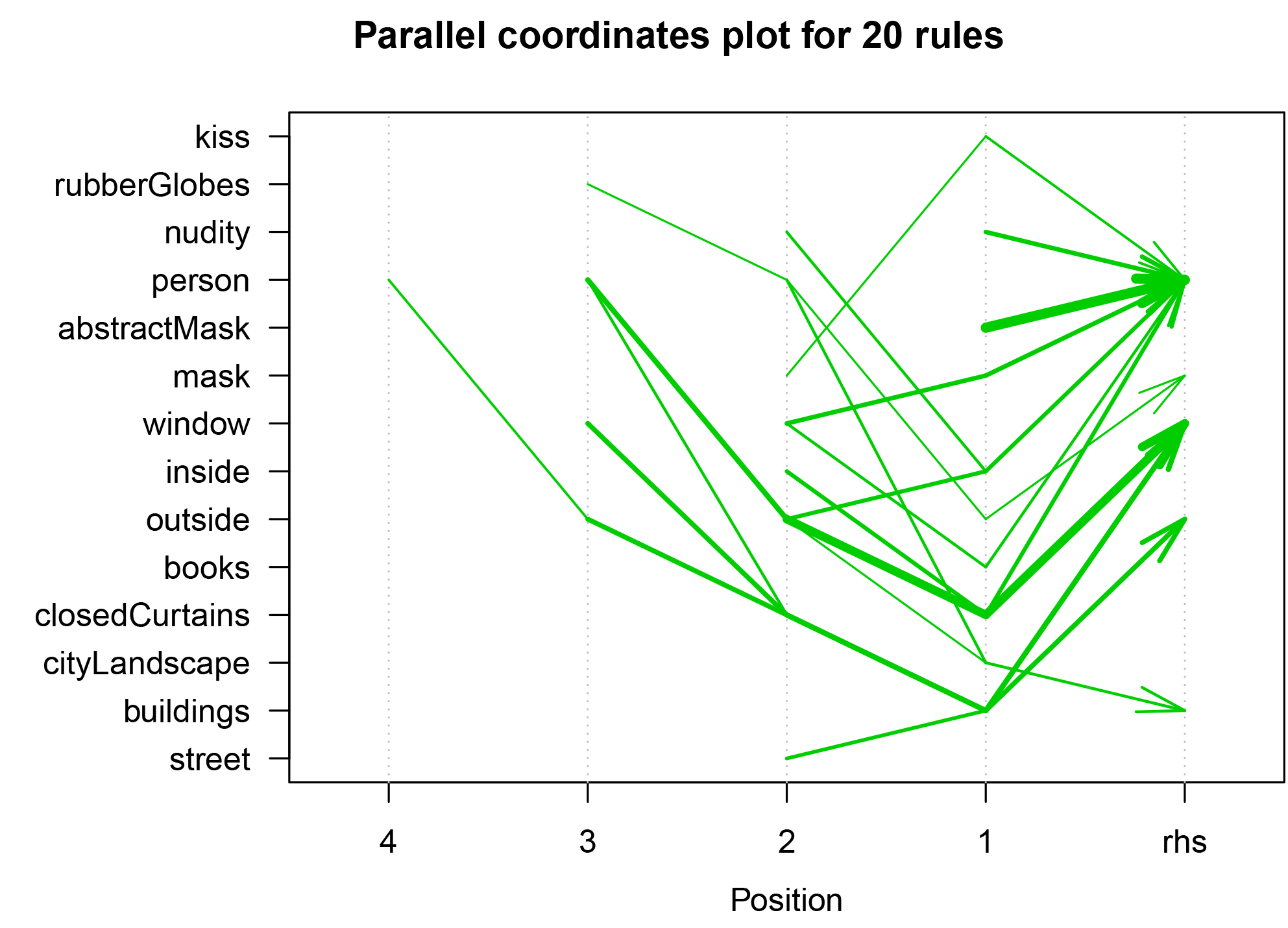}
	\caption{Parallel representation graph (paracoord) identifying the main relationships that connect with RHS}
	\label{fig:fig7}
\end{figure}

Figure \ref{fig:fig5} below graphically shows the map of association relationships between objects in the same image from sample established in Table 1, where the radius of the circle represents the support of the measure and the intensity of color. The four elements represented with the most antecedent and consequent rules of association in the sample are: a) Person, b) Window, c) Building, d) Outside. (See Figures \ref{fig:fig6} and \ref{fig:fig7}).

\section{Discussion}
Connecting with the main idea of how visual representation is based on the selection, classification and decontextualization of elements to forge a meaning through the way they are structured, computer analysis has revealed the elements most used and their associations.

We have been able to see that, through quantification in terms of the frequency of the objects contained in the images, the isolated elements most commonly used by creators for their visual representations were people, an outside view, an interior scene, face masks, windows, vegetable elements and buildings. In this regard, it is worth highlighting the high frequency with which the object "face mask" appears, how a foreign element in our lives appears in representations associated with the pandemic and has incorporated itself into collective imagery.

However, the frequency study does not allow one to know the use of the elements and their relationships in creating holistic discourses on the Pandemic. For this, a methodology of graphic representation of rules of association, which shows us that the use of people, windows, buildings, and outside scenes are the elements that have related the most with others to construct the visual representation of subjective experiences in the context of the COVID-19 pandemic.
We did not find images of saturated hospitals, deaths, rising poverty or increased social tension, or they are not representative in the sample

\section{Conclusions}
The study of the denotative elements that make up the images produced in the context of the COVID-19 pandemic and obtained from The Covid Art Museum account on the social network Instagram has helped us to understand how we associate a series of objects and images on a concrete reality, establishing a visual representation of the same. 

Thus, we can see how Internet users have shared creations that are testimonies to their experiences with a global event. But, despite the individualism of the discourses and creations, the representations share certain patterns and ideas when it comes to choosing the denotative elements, where the human figure focuses the discourse and the window element is detected as the link with confinement, the restriction most emphasized by creators as a symbol of the limitation of movement, in addition to representations of buildings, the habitat where the experience has taken place, and the representation of outdoor spaces as a representation of the desire to regain freedom. In this way, one can see that the construction of visual narratives among the images, in terms of associations, is produced by means of these four denotative elements (people, buildings, windows and outdoor scenes) and this has generated visual discourses, not so much on the disease, but its impact on everyday life. 

In the sample, no significant uses of symbolic or iconographic elements traditionally associated with disease were detected (e.g., a skull as a reference to death, the prostrate and weak individual, the clock as a reference to the time that has passed, etc.), iconographic features that have even been analyzed in relation to previous pandemics in the works of Ana María Diaz Pérez (1991)  or, more recently, that of Carla Joan Dias de Angelo (\citep{carla_muerte_2018}).

The results suggest that the experience of the disease has focused on the collateral damage that has been caused and the changes in our everyday dynamics, beyond the physical, economic, and social damage, etc. 
Based on the results, we can conclude that there are a series of shared patterns and images that are repeated, each one originating in different authors, thus revealing a shared collective imagery, to a certain extent inherent to a globalized society that has experienced a common pandemic, and the use of social networks as a global channel of expression, which have constituted a visual imagery but with limited use of resources and codes in creations

\section{Limitations and Future research}
Recent research by Elena Semino (\citeyear{semino_covid-19_2020}) on the metaphors have been generated at the narrative level around the pandemic establishes that: «Metaphors are crucial tools for communication and thinking, and can be particularly useful in public health communication. For example, the 'second wave' metaphor suggests that there is renewed danger and threat from the virus, and may therefore encourage compliance with measures aimed at reducing transmission» (p.5). 

This metaphor is a representative example of how the pandemic has been described at the narrative level and how through the metaphor it is possible to build a concrete reality. In this sense, it is interesting to note how at the visual level, from a freer and more personal position like artistic discourse, metaphors have been built in a very different way from the way done in verbal discourse. Thus, 'waves' or 'battlefields'—narrative metaphors described in Semino (\citeyear{semino_covid-19_2020})—give way to other narratives, which comprise a field to be explored in future research.

The methodology proposed in our research has allowed relations between the objects that underlay individual representations to be empirically established. Based on the aforementioned conclusions, there is significant room for work to extract the vast amount of information contained in images, such as the color, shape, spatial location of elements, among others, and in that way study the process that leads artists to define the elements contained in a work.

Regarding the YOLO tool, which distinguishes at least 80 categories natively, our research has proposed detecting 64 categories other than those already defined by YOLO, taking the network structure and training weights as the basis. We will consider reducing this number in future improvements, grouping similar concepts together, thus facilitating the search for shared patterns among the different artistic representations and, in particular, expanding the characteristics extracted from the images that allow us to find other new relations on the object categories.

From a more netnographic approach, the creators’ demographic details could be used to analyze whether socially and culturally contextualized representations are produced or whether they are globalized images using codes associated with the pandemic instead.

Thus, this qualitative analysis will be able to define what types of social representations have been established regarding the pandemic, alluding to the Moscovici \citeyear{moscovic_introduccion_1985} classification and in that way determine the possible contributions to Health Humanities

\bibliographystyle{unsrtnat}


\end{document}